\documentclass[10pt,twocolumn,letterpaper]{article}

\usepackage{cvpr}              

\newcommand{\mypara}[1]{\vspace{2pt}\noindent{\bf{#1}}}
\usepackage[dvipsnames]{xcolor}

\usepackage{listings}
\usepackage{pxfonts}
\usepackage{adjustbox}
\usepackage{svg}

\usepackage{xcolor}
\definecolor{codegreen}{rgb}{0,0.6,0}
\definecolor{codegray}{rgb}{0.5,0.5,0.5}
\definecolor{codepurple}{rgb}{0.58,0,0.82}
\definecolor{backcolour}{rgb}{0.95,0.95,0.92}

\lstdefinestyle{mystyle}{
    backgroundcolor=\color{backcolour},   
    commentstyle=\color{codegreen},
    keywordstyle=\color{magenta},
    numberstyle=\tiny\color{codegray},
    stringstyle=\color{codepurple},
    basicstyle=\ttfamily\footnotesize,
    breakatwhitespace=false,         
    breaklines=true,                 
    captionpos=b,                    
    keepspaces=true,                 
    numbers=left,                    
    numbersep=5pt,                  
    showspaces=false,                
    showstringspaces=false,
    showtabs=false,                  
    tabsize=2
}

\definecolor{cvprblue}{rgb}{0.21,0.49,0.74}
\usepackage[pagebackref,breaklinks,colorlinks,citecolor=cvprblue]{hyperref}

\def\paperID{12} 
\def\confName{CVPR}
\def\confYear{2024}

\title{Audio-Visual Generalized Zero-Shot Learning using\\ Pre-Trained Large Multi-Modal Models}

\author{David Kurzendörfer\hspace{1pt}\thanks{Denotes equal contribution}\hspace{3pt}\textsuperscript{1,2}, \hspace{2pt} Otniel-Bogdan Mercea$^*$\textsuperscript{1,3,4}, \hspace{2pt} A. Sophia Koepke\textsuperscript{1,3}, \hspace{2pt} 
Zeynep Akata\textsuperscript{3,4,5}  \\ \\
\textsuperscript{1}University of T{\"u}bingen \hspace{2pt}
{\textsuperscript{2}Localyzer GmbH \hspace{2pt}
\textsuperscript{3}Tübingen AI Center \hspace{2pt}
\textsuperscript{4}Helmholtz Munich}\\
{\textsuperscript{5}Technical University of Munich} \\
{\small \tt \{otniel-bogdan.mercea, a-sophia.koepke\}@uni-tuebingen.de} \\
{\small \tt dk@localyzer.de}, \quad {\small \tt zeynep.akata@helmholtz-munich.de}
}

\begin{document}
\maketitle
\begin{abstract}

\noindent Audio-visual zero-shot learning methods commonly build on features extracted from pre-trained models, e.g.\ video or audio classification models. However, existing benchmarks predate the popularization of large multi-modal models, such as CLIP and CLAP. In this work, we explore such large pre-trained models to obtain features, i.e.\ CLIP for visual features, and CLAP for audio features. Furthermore, the CLIP and CLAP text encoders provide class label embeddings which are combined to boost the performance of the system. 
We propose a simple yet effective model that only relies on feed-forward neural networks, exploiting the strong generalization capabilities of the new audio, visual and textual features. Our framework achieves state-of-the-art performance on $\text{VGGSound-GZSL}^{cls}$, $\text{UCF-GZSL}^{cls}$, and $\text{ActivityNet-GZSL}^{cls}$ with our new features. Code and data available at: \url{https://github.com/dkurzend/ClipClap-GZSL}.
\end{abstract}
  
\section{Introduction}
\label{sec:intro}

\begin{figure}[!t]
  \centering
  \includegraphics[width=0.5\textwidth]{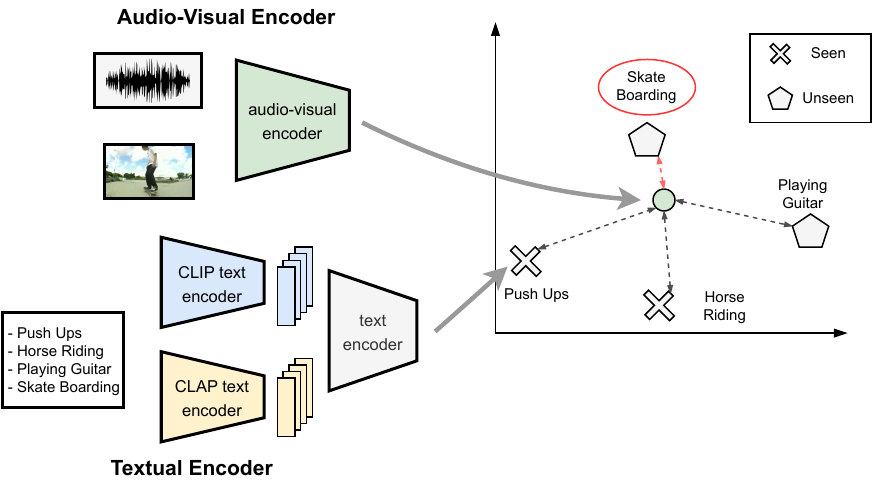}
  \caption{Our framework for audio-visual GZSL maps the audio and visual data to embeddings that are aligned with class label embeddings that are obtained from merging CLIP and CLAP embeddings. The class label embedding that is closest to the audio-visual embedding determines the class prediction. At test time, the set of class label embeddings contains both seen and unseen classes. }
  \label{fig:av-gzsl}
  \vspace{-1.0em}
\end{figure}

The synergy of audio and visual modalities is a valuable asset for tasks like video classification. Imagine a bustling street captured on camera, where the integration of audio—such as footsteps, car engines, or a dog barking—provides crucial context for interpreting the visual content. In practical deep learning applications, models often encounter new and unseen data, e.g.\ objects or scenes not present in their training data. This challenge arises due to the vast diversity of real-world data and the impracticality of preparing models for every possible variation.  
A well-designed deep learning model should exhibit the ability to transfer knowledge from familiar classes to unseen ones.

Audio-visual generalized zero-shot learning (GZSL) aims at classifying videos using audio and visual inputs. Previous work~\cite{cjme, avgzslnet, AVCA, TCAF, hyper_avca} learns to align audio-visual embeddings with corresponding class label embeddings. The class label embedding that is closest to the audio-visual embedding is chosen for the prediction. Existing audio-visual GZSL methods build on features obtained from pre-trained models for the audio, visual and textual data. 
However, the feature extraction methods \cite{vggish, C3D, selavi, carreira2017quo, kumar2018knowledge} used in most previous works \cite{TCAF, hyper_avca, cjme, avgzslnet} do not reflect the state of the art anymore. 
In recent years, the transformer architecture \cite{attention} has proved successful in many areas such as natural language processing \cite{CLIP, BERT, gpt4}, the vision domain \cite{ViT, CLIP} or the audio domain \cite{audio_spectrogram_transformer, HTSAT}.  
CLIP \cite{CLIP} is a popular vision-language model which contains transformers as the text and image encoders that map to a joint multi-modal embedding space.  
\cite{wavcaps} introduces CLAP, a similar method for the audio-language domain.

In this work, we address audio-visual GZSL by exploring pre-trained multi-modal models to produce audio, visual, and textual input features. We show that the high generalization capabilities of such large pre-trained models are beneficial in the GZSL setting.
We use CLIP~\cite{CLIP} for visual feature extraction and CLAP~\cite{wavcaps} for audio feature extraction. Both models contain text encoders which provide input class label embeddings. Consequently, a novel feature of our method compared to prior work (e.g.\ \cite{cjme, avgzslnet, AVCA, TCAF, hyper_avca}) is the usage of two class label embeddings that are aggregated into a unified label embedding. Since the textual embeddings are obtained from vision-langauge and audio-language models, the audio and visual input features are already aligned with the corresponding class embeddings. 
Our proposed model (see Figure~\ref{fig:av-gzsl} for an overview) ingests the aforementioned input features and class label embeddings, only relying on simple feed-forward neural networks in conjunction with a composite loss function.  
Our contributions can be summarized as follows: 

\begin{itemize}
  \item Our proposed audio-visual GZSL framework builds on features from pre-trained multi-modal models. Moreover, we exploit the text encoders in the same multi-modal models to provide two class label embeddings that are combined to form a unified and robust textual class label embedding;
  \item Our simple but effective framework achieves state-of-the-art results on the VGGSound-GZSL$^{cls}$, UCF-GZSL$^{cls}$ and ActivityNet-GZSL$^{cls}$ datasets when using the new input features;
  \item Qualitative analysis shows that our approach produces well-separated clusters for the seen and unseen classes in the embedding space.
\end{itemize}

\section{Related Work}
\label{sec:related_work}

In this section, we summarize related work concerned with audio-visual learning, zero-shot learning, and audio-visual generalized zero-shot learning.

\mypara{Audio-Visual Learning.} Using audio data for video analysis can significantly enhance the visual representations, for instance for sound source localization \cite{afouras2022self, arandjelovic2018objects, chen2021localizing,qian2020multiple, tian2018audio, xu2020cross}, sound source separation \cite{gao2019co, tzinis2020into,zhu2022v} or both sound source localisation and separation \cite{afouras2020self, owens2018audio, zhao2018sound, zhao2019sound}.
Various works perform the audio-visual correspondence task \cite{afouras2022self, arandjelovic2018objects, chen2021localizing, qian2020multiple, owens2016ambient, aytar2016soundnet,owens2018learning} or synchronization task \cite{owens2018audio, zhao2018sound, chen2021audio, chung2017out, ebeneze2021detection, khosravan2019attention,cheng2020look, korbar2018cooperative,xiao2020audiovisual,korbar2018cooperative} to learn representations that contain useful knowledge of both modalities. 
Moreover, \cite{alwassel2020self, selavi, chen2021distilling, cheng2020look,  nagrani2021attention, patrick2020multi, xiao2020audiovisual, mercea2023avdiff} 
learn rich audio-visual representations. For example, \cite{alwassel2020self, selavi, cheng2020look, patrick2020multi, xiao2020audiovisual} use self-supervision to learn these rich audio-visual representations, while \cite{chen2021distilling} uses knowledge distillation, and \cite{nagrani2021attention,mercea2023avdiff} use a supervised learning objective along with a transformer specifically designed to merge the audio and the visual modalities.
Multiple works combine the audio and visual modalities for speech recognition and lip reading \cite{afouras2018deep, afouras2020asr, nagrani2020disentangled, ma2021end}. Other tasks where audio and visual modalities are combined include spotting of spoken keywords \cite{momeni2020seeing} and audio synthesis from visual information \cite{gan2020foley, goldstein2018guitar, koepke2020sight, koepke2019visual, narasimhan2022strumming, su2020multi, su2021does, zhou2019vision}. 

\mypara{Zero-Shot Learning (ZSL)} involves training a model to classify new test classes not seen during training, e.g.\ by learning a mapping between input features and semantic embeddings. Typically, the semantic embeddings are obtained as text embeddings from class labels~\cite{akata2015evaluation, xian2018feature, frome2013devise, norouzi2013zero,xian2016latent, akata2015label} and from class attributes and descriptions \cite{xian2018feature, lampert2013attribute, romera2015embarrassingly}.
Other works use generative methods to synthesize data for unseen classes \cite{xian2018feature, verma2018generalized, hao2023contrastive, gupta2023generative, gowda2023synthetic}. 
In \cite{CLIP, clap, wavcaps, li2017learning, pham2111combined, yu2022coca, zhu2022uni}, ZSL is performed by applying a pre-trained model on new, unseen datasets.
Recently, CLIP text and image encoders have been integrated into various ZSL frameworks~\cite{haas2023learning, wu2022im2city, zhou2023zegclip, ding2022decoupling, xu2022simple, luo2023segclip, liang2023open, li2023rs, novack2023chils, wang2023clipn}, e.g. for zero-shot / open-vocabulary semantic segmentation \cite{zhou2023zegclip, ding2022decoupling, xu2022simple, luo2023segclip, liang2023open}.
Taking advantage of the strong generalization ability of large pretrained multi-modal models, we use CLIP~\cite{CLIP} and CLAP~\cite{wavcaps} as feature extraction methods for the visual and audio domain.

\mypara{Audio-visual GZSL} was first introduced in~\cite{cjme} which proposed the AudioSetZSL datset. \cite{cjme,avgzslnet} both proposed methods that map the audio, visual, and textual input features (i.e.\ word2vec~\cite{word2vec}) to a joint embedding space. 
\cite{AVCA} curated several new benchmarks for the audio-visual GZSL task, along with a framework that uses cross-attention between the audio and the visual modalities.  
\cite{hyper_avca} additionally uses a hyperbolic alignment loss. 
While \cite{cjme, avgzslnet, AVCA, hyper_avca} ingest temporally averaged audio and visual features from pre-trained audio and visual classifiers, \cite{TCAF} exploits the inherent temporal structure of videos. 
In contrast to prior work that fuses the audio and visual information at later stages, our method directly concatenates audio and visual input features before passing them into a feed-forward neural network.
Furthermore, our proposed method utilizes two input class label embeddings obtained from CLIP and CLAP.

\section{Proposed Approach}
\label{sec:method}
In this section, we motivate the use of features extracted from large pre-trained multi-modal models, describe the audio-visual GZSL setting and our proposed framework and training objective.

The audio-visual GZSL benchmarks introduced in prior work~\cite{AVCA,TCAF} build on features extracted from audio and video classification networks. However, those feature extraction methods date back to 2017 and 2015 for the audio~\cite{vggish} and visual features~\cite{C3D} respectively. CLIP~\cite{CLIP} and CLAP~\cite{wavcaps} have shown impressive generalization capabilities. 
We propose to use features extracted from CLIP and CLAP as inputs to our framework, eliminating the need for a complex architecture to adapt to the audio-visual GZSL task.
We use text embeddings obtained from CLIP and CLAP which are aligned with corresponding audio / visual features. 

\mypara{Audio-visual GZSL setting.} 
In the ZSL setting, two disjoint sets of classes are considered, i.e. seen and unseen classes $\text{S}$ and $\text{U}$ with $\text{S} \cap \text{U}=\varnothing$. 
In ZSL, the model is trained on the seen classes and later evaluated on the test set, which only consists of unseen classes.   
In the GZSL setting the model is trained on seen classes ($\text{S}$), but the test set contains both seen and unseen classes, making this scenario more realistic. 

Formally, the set of data samples that belong to the seen classes is denoted by $\text{S}=\left(v_i^s, a_i^s, w_i^s, y_{i}^s\right)_{i \in\{1, \cdots, N\}}$ where each data point $i$ is a quadruple where $a_i^s$ is the audio feature, $v_i^s$ is the visual feature, $y_{i}^s$ is the ground-truth class label of sample $i$ and $w_i^s$ is the textual label embedding corresponding to the ground-truth class label. $N$ is the number of samples in $S$.
Likewise, the set of samples from unseen classes of size $M$ is defined as $\text{U}=\left(v_i^u, a_i^u, w_i^u, y_{i}^u\right)_{i \in\{1, \cdots, M\}}$. In GZSL, the goal is to learn a function $h:\left(v_i^s, a_i^s\right) \mapsto w_i^s$ which for samples from unseen classes fulfills $h\left(v_i^u, a_i^u\right)=w_i^u$. The total number of classes is denoted as $K$ and the class label $j \in \{1, \ ,2 \ \dots, \ K \}$. The number of seen and unseen classes are denoted as $K_s$ and $K_u$.

\mypara{Model architecture.}
\begin{figure}[!t]
  \centering
  \includegraphics[width=0.5\textwidth]{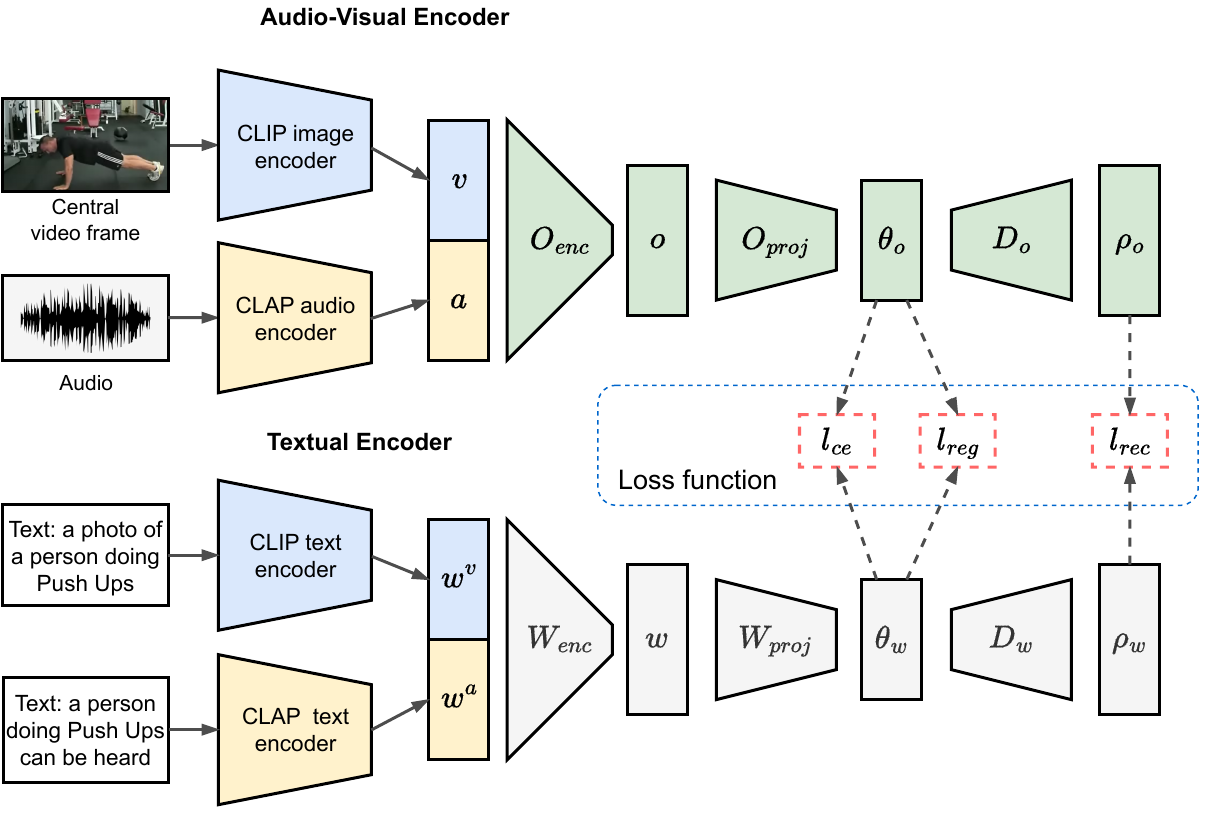}
  \caption{The image and audio encoders of CLIP and CLAP are used to extract features from the raw input which are concatenated and passed through multiple feed-forward networks to get an audio-visual output embedding $\theta_o$. Likewise, the text encoders of CLIP and CLAP are used to extract textual label embeddings. They are passed through a series of neural networks to obtain a learned class label embedding $\theta_w$. Both $\theta_o$ and $\theta_w$ reside in a joint embedding space.}
  \label{fig:model_architecture}
  \vspace{-1.0em}
\end{figure}
Our proposed model is visualized in \cref{fig:model_architecture}. 
It accepts audio and visual features as inputs, denoted as $a \in \mathbb{R}^{d_{in_{a}}}$ and $v \in \mathbb{R}^{d_{in_{v}}}$ respectively (for simplicity, subscripts $i$ denoting the $i^{th}$ sample are dropped). These features are obtained using CLAP and CLIP as feature extractors. 
In addition, our model takes as input text embeddings $w^v \in 
\mathbb{R}^{d_{in_{v}}}$ from CLIP and $w^a \in \mathbb{R}^{d_{in_{a}}}$ from CLAP. CLIP and CLAP merely serve as feature extractors and thus are not optimized when training our framework.
Our proposed model consists of a branch for the audio-visual features, and a branch for the textual label embeddings.
In the audio-visual branch, the inputs $a$ and $v$ are first concatenated and then passed through an encoder block $O_{enc}$ to produce the audio-visual input
\begin{equation}
    o = O_{enc}(\operatorname{concat}\left(  v, a\right)) ,
\end{equation}
where $o \in \mathbb{R}^{d_{model}}$. $O_{enc}$ consists of a linear layer $f_{O_{enc}}: \mathbb{R}^{(d_{in_{v}} + d_{in_{a}})} \rightarrow \mathbb{R}^{d_{model}}$, followed by batch normalization \cite{batchnorm}, a ReLU activation function \cite{relu}, and dropout~\cite{dropout}.
To get the final audio-visual embedding $\theta_o \in \mathbb{R}^{d_{out}}$ that is used for the prediction, $o$ is passed through a projection network
$\theta_o = O_{proj}(o)$, where $\theta_o$ is
composed of linear layers $f_{O_{proj}}^1: \mathbb{R}^{d_{model}} \rightarrow \mathbb{R}^{d_{hidden}}$ and $f_{O_{proj}}^2: \mathbb{R}^{d_{hidden}} \rightarrow \mathbb{R}^{d_{out}}$. Both layers are followed by batch normalization, ReLU, and dropout.

The textual branch follows a similar structure as the audio-visual branch. First, $w^a$ and $w^v$ are concatenated and are input into an encoder network to generate a unified text embedding $w \in \mathbb{R}^{d_{model}}$,
\begin{equation}
    w = W_{enc}(\operatorname{concat}\left(  w^v, w^a\right)).
\end{equation}
$W_{enc}$ contains a linear layer $f_{W_{enc}}: \mathbb{R}^{(d_{in_{v}} + d_{in_{a}})} \rightarrow \mathbb{R}^{d_{model}}$ followed by batch normalization, ReLU, and dropout. The output $w$ is further processed by a projection layer $\theta_w = W_{proj}(w)$,
where $\theta_w \in \mathbb{R}^{d_{out}}$. $ W_{proj}$ is given by a linear layer $f_{W_{proj}}: \mathbb{R}^{d_{model}} \rightarrow \mathbb{R}^{d_{out}}$ with batch normalization, ReLU, and dropout. The goal of the model is to align the projected label embedding $\theta_w$ and the audio-visual output embedding $\theta_o$ in a joint embedding space of dimension $d_{out}$, such that $\theta_o$ is closest to the $\theta_w$ that corresponds to the ground-truth class. 

At test time, classification is done by calculating $\theta_w$ for all the classes and determining the class label $c$ that is closest to $\theta_o$:
\begin{equation}
\label{eq:classification_rule}
c=\underset{j}{\operatorname{argmin}}\left(\left\|\theta_{w}^j-\theta_o\right\|_2\right),
\end{equation}
where $\theta_{w}^j$ is the output label embedding $\theta_{w}$ for class $j$.

\mypara{Training objective.}
The loss function $l$ used to train our framework is adopted from \cite{TCAF} and consists of a cross-entropy loss $l_{ce}$, a reconstruction loss $l_{rec}$, and a regression loss $l_{reg}$. The final loss is then given by
\begin{equation}
l=l_{ce}+l_{rec}+l_{reg}.
\end{equation}

The \textbf{cross-entropy loss} is given by 
\begin{equation}
l_{ce}=-\frac{1}{n} \sum_i^n \log \left(\frac{\exp \left(\theta_{w_{seen}, k_{gt_{i}}} \theta_{o_i}\right)}{\sum_{k_j}^{K_{s}} \exp \left(\theta_{w_{seen}, k_j} \theta_{o_i}\right)}\right),
\end{equation}
where $\theta_{w_{seen}} \in \mathbb{R}^{K_{s} \times d_{out}}$ denotes the matrix of the projected class embeddings for the seen classes. $k_{gt} \in \{1, \ ,2 \ \dots, \ K_s \}$ refers to the ground-truth class index, and thus $\theta_{w_{seen}, k_{gt_{i}}}$ selects the row of $\theta_{w_{seen}}$ that belongs to the target of the current sample $i$. The number of training samples is denoted by $n$. 

The \textbf{reconstruction loss} 
semantically aligns the output embeddings $\theta_o$ and $\theta_w$. For this, two decoder networks, $D_o$ and $D_w$, are used to obtain $\rho_o = D_{o}(\theta_o)$ with $\rho_o \in \mathbb{R}^{d_{model}}$. 
$D_o$ consists of two linear layers $f_{D_{o}}^1: \mathbb{R}^{d_{out}} \rightarrow \mathbb{R}^{d_{hidden}}$ and  $f_{D_{o}}^2: \mathbb{R}^{d_{hidden}} \rightarrow \mathbb{R}^{d_{model}}$ which are both followed by batch normalization, ReLU, and dropout.
$D_w$ gets $\theta_w$ as input, such that
$\rho_w = D_{w}(\theta_w)$,
where $\rho_w \in \mathbb{R}^{d_{model}}$. $D_w$ consists of one linear layer $f_{D_{w}}: \mathbb{R}^{d_{out}} \rightarrow \mathbb{R}^{d_{model}}$ with batch normalization, ReLU, and dropout.

The reconstruction loss encourages the reconstructions $\rho_o$ and $\rho_w$ to be close to the label embedding $w$ by minimizing 
\begin{equation}
l_{rec}=\frac{1}{n} \sum_{i=1}^n\left(\rho_{o_i}-w_i\right)^2+\frac{1}{n} \sum_{i=1}^n\left(\rho_{w_i}-w_i\right)^2,
\end{equation}
where n is the number of training samples.

The \textbf{regression loss} computes the mean squared error between the output embeddings of the model and the ground-truth label embeddings:
\begin{equation}
l_{reg}=\frac{1}{n} \sum_{i=1}^n\left(\theta_{o_i}-\theta_{w_i}\right)^2,
\end{equation}
where $\theta_{o_i}$ is the audio-visual embedding for sample $i$ and $\theta_{w_i}$ is the corresponding output label embedding.

\section{Experiments}
\label{sec:experiments}

In this section, we describe our experimental setup (\cref{subsec:experimental_setup}), our results (\cref{subsec:experimental_results}), and ablate crucial components of our framework (\cref{subsec:ablations}).

\subsection{Experimental setup}\label{subsec:experimental_setup}
Here, we describe the evaluation metrics, the features used, implementation details, and our baselines.

\mypara{Evaluation metrics.} We evaluate our audio-visual GZSL method on the $\text{VGGSound-GZSL}^{cls}$, $\text{UCF-GZSL}^{cls}$, and $\text{ActivityNet-GZSL}^{cls}$ datasets introduced in \cite{AVCA} and as suggested in \cite{TCAF} (instead of using the $\textit{main split}$ also introduced in \cite{AVCA}). 

We follow \cite{zsl_good_bad_ugly, AVCA, TCAF} and report the mean class accuracy scores for the seen classes ($acc_{\text{S}}$) and unseen classes ($acc_{\text{U}}$) separately. For the GZSL performance metric, their harmonic mean is obtained as
\begin{equation}
H M=\frac{2 * acc_{\text{U}} * acc_{\text{S}}}{acc_{\text{U}}+acc_{\text{S}}}.
\end{equation}
In addition, we calculate the zero-shot learning performance as the mean class accuracy $acc_{\text{ZSL}}$ for the unseen classes. In this setting, only classes from the subset of unseen test classes
can be selected as prediction.

\mypara{Feature extraction.}
We do not rely on the same feature extractors as previous work \cite{AVCA, TCAF, hyper_avca}. Instead, visual features $v_i$ and class label embeddings $w_j^v$ are extracted from the videos using \textbf{CLIP}~\cite{CLIP}.  
For each video, the middle frame is passed through the image encoder of ViT-B/32 CLIP model, yielding a $512-$dimensional feature vector. 
In addition, for each class label, a $512-$dimensional textual embedding is extracted using the CLIP text encoder. Here, we follow \cite{CLIP}, which recommends the usage of text prompt ensembles. We provide more details and a concrete list of text prompts in the supplementary materials in \ref{supplement:clip_feature_extraction}.

Likewise, audio features $a_i$ and class label embeddings $w_j^{a}$ are extracted using \textbf{CLAP}~\cite{wavcaps}.

The raw audio data is resampled to $32000$ Hz and center cropped or zero-padded to $10$ seconds, depending on the audio length. 64-dimensional log mel-spectrograms are extracted from the audio by using a 1024-point Hanning window with a hop size of $320$.
 Audio embeddings are obtained from the audio encoder of CLAP, and text embeddings from its ext encoder. The joint audio and textual embedding space in CLAP is of size $1024$. For the class text embeddings, text prompt ensembles are used similar to CLIP (See more details in the supplementary materials \ref{supplement:clap_feature_extraction}).

\mypara{Implementation details.}
To train our framework, we use the Adam optimizer \cite{adam_optimizer} with weight decay $1e^{-5}$, $\beta_1 = 0.9$ and $\beta_2 = 0.999$, and a batch size of $64$. Furthermore, the initial learning rates are $1e^{-4}$ / $7e^{-5}$ / $1e^{-4}$ for $\text{VGGSound-GZSL}^{cls}$ / $\text{UCF-GZSL}^{cls}$ / and $\text{ActivityNet-GZSL}^{cls}$. When the validation HM score does not improve for 3 consecutive epochs during training, the learning rate is reduced by a factor of $0.1$. 
In the first training stage, the models for $\text{VGGSound-GZSL}^{cls}$ and $\text{ActivityNet-GZSL}^{cls}$ were trained for $15$ epochs, while for $\text{UCF-GZSL}^{cls}$ we used $20$ epochs. 
Calibrated stacking~\cite{calibrated_stacking} is a scalar which biases the output of the network towards unseen classes, as the network without the calibration is significantly biased towards seen classes. To address the inherent bias of ZSL methods towards seen classes, we use calibrated stacking with search space interval [0, 5] and a step size of $0.07$. 

We follow the training and evaluation protocol of \cite{AVCA, TCAF, hyper_avca} for our method. The training is divided into two stages. In the first stage, models are trained on the training set. The validation set comprising val(U) and val(S) is used to determine model parameters such as those for calibrated stacking and the best epoch based on the HM score. 
In the second training stage, the models are trained again using the parameters determined in the first stage, however, this time, the models are trained on the union of the training and validation set \{train $\cup$ val(S) $\cup$ val(U)\}. Finally, the final results on the test set are obtained by evaluating the models trained in the second stage.

The inputs are of size $d_{in_{a}} = 1024$ and $d_{in_{v}} = 512$. For the $\text{VGGSound-GZSL}^{cls}$, $\text{UCF-GZSL}^{cls}$ and $\text{ActivityNet-GZSL}^{cls}$ datasets, the model dimension $d_{model} = 512$ is chosen, and the output dimension is set to $d_{out} = 64$. In addition, for all three datasets, $d_{hidden}$ is set to 512, and the dropout rate is set to $0.1$.
All models were trained on a single NVIDIA GeForce RTX 2080 Ti GPU.

\mypara{Baselines.}
We compare our framework with the state-of-the-art methods CJME \cite{cjme}, AVGZSLNet \cite{avgzslnet}, AVCA \cite{AVCA}, and Hyper-multiple \cite{hyper_avca}. For CJME, AVGZSLNet and AVCA we use the training parameters from \cite{AVCA} and we evaluate Hyper-multiple using the training parameters from \cite{hyper_avca}. All methods are evaluated using the new CLIP and CLAP features. For fairness, we adjust the baseline methods for our input representation to using two textual input embeddings, by appending an additional layer at the beginning of the network as
\begin{equation}
    w_i = W_{enc}(\operatorname{concat}\left(  w_i^v, w_i^a\right)),
\end{equation}
where $w_i^v \in \mathbb{R}^{512}$ is the CLIP class label embedding for sample $i$ and $w_i^a \in \mathbb{R}^{1024}$ is the CLAP class label embedding.

\setlength{\tabcolsep}{8pt} 
\renewcommand{\arraystretch}{1.1}
\begin{table*}[tp]
\centering
\begin{adjustbox}{width=\textwidth}
\begin{tabular}{l|cccc|cccc|cccc}
\hline
Method     & \multicolumn{4}{c|}{$\text{VGGSound-GZSL}^{cls}$} & \multicolumn{4}{c|}{$\text{UCF-GZSL}^{cls}$}      & \multicolumn{4}{c}{$\text{ActivityNet-GZSL}^{cls}$}        \\
           & $acc_{\text{S}}$      & $acc_{\text{U}}$     & HM    & $acc_{\text{ZSL}}$  & $acc_{\text{S}}$     & $acc_{\text{U}}$     & HM    & $acc_{\text{ZSL}}$   & $acc_{\text{S}}$     & $acc_{\text{U}}$              & HM    & $acc_{\text{ZSL}}$   \\ \hline
CJME~\cite{cjme}       & 11.96  & 5.41  & 7.45  & 6.84 & 48.18 & 17.68 & 25.87 & 20.46 & 16.06 & 9.13           & 11.64 & 9.92  \\
AVGZSLNet~\cite{avgzslnet}  & 13.02  & 2.88  & 4.71  & 5.44 & 56.26 & 34.37 & 42.67 & 35.66 & 14.81 & 11.11          & 12.70 & 12.39 \\
AVCA~\cite{AVCA}       & \textbf{32.47}  & 6.81  & 11.26 & 8.16 & 34.90 & 38.67 & 36.69 & 38.67 & 24.04 & 19.88          & 21.76 & 20.88 \\
Hyper-multiple~\cite{hyper_avca} & 21.99  & 8.12  & 11.87  & 8.47 & 43.52 & 39.77 & 41.56 & 40.28 & 20.52 & \textbf{21.30} & 20.90 & 22.18 \\ \hline
Ours &
  29.68 &
  \textbf{11.12} &
  \textbf{16.18} &
  \textbf{11.53} &
  \textbf{77.14} &
  \textbf{43.91} &
  \textbf{55.97} &
  \textbf{46.96} &
  \textbf{45.98} &
  20.06 &
  \textbf{27.93} &
  \textbf{22.76} \\ \hline
\end{tabular}
\end{adjustbox}
\caption{Performance of our model compared to state-of-the-art methods for audio-visual (G)ZSL on the $\text{VGGSound-GZSL}^{cls}$, $\text{UCF-GZSL}^{cls}$ and $\text{ActivityNet-GZSL}^{cls}$ datasets. For a fair comparison, all baselines are also trained and evaluated using both CLIP and CLAP features and class label embeddings. We report the mean class accuracy for seen ($acc_{\text{S}}$) and unseen ($acc_{\text{U}}$) classes, along with their harmonic mean (HM) for GZSL performance. In addition, ZSL performance ($acc_{\text{ZSL}}$) is reported.}
\label{tab:main_results}
\vspace{-1.0em}
\end{table*}

\subsection{Experimental results}\label{subsec:experimental_results}
In this section, we present quantitative and qualitative results for audio-visual GZSL obtained with our proposed framework.

\mypara{Quantitative results.} On all three datasets, our model outperforms all baseline methods in terms of GZSL performance (HM) as can be seen in \cref{tab:main_results}. For example, on $\text{UCF-GZSL}^{cls}$, we achieve a HM score of $55.97\%$ whereas the next best baseline (AVGZSLNet) achieves $42.67\%$. Similarly, on $\text{ActivityNet-GZSL}^{cls}$, our model achieves a HM score of $27.93\%$ while Hyer-multiple achieves $20.90\%$, showing an improvement of $7.03\%$. Finally, on $\text{VGGSound-GZSL}^{cls}$, our model achieves $16.18\%$, compared to $11.87\%$ for Hyper-multiple.

Our method also outperforms all the baselines on all three datasets for ZSL ($acc_{\text{ZSL}}$). On $\text{VGGSound-GZSL}^{cls}$, we achieve a $acc_{\text{ZSL}}$ score of $11.53\%$ while the second-best method achieves a score of $8.47\%$. On $\text{UCF-GZSL}^{cls}$, our method achieves a $acc_{\text{ZSL}}$ performance of $46.96\%$, compared to $40.28\%$ for Hyper-multiple. On $\text{ActivityNet-GZSL}^{cls}$, the difference in ZSL performance is very small. Our model obtains a $acc_{\text{ZSL}}$ score of $22.76\%$ while Hyper-multiple achieves $22.18\%$. 

In terms of the seen and unseen scores $acc_{S}$ and $acc_{U}$, our method is the best performing model most of the time. On $\text{VGGSound-GZSL}^{cls}$ we achieve the second best $acc_{S}$ of $29.68\%$, while AVCA achieves $32.47\%$. For the $acc_{U}$, our model performs best with $11.12\%$ vs. $8.12\%$ achieved by Hyper-multiple. On $\text{UCF-GZSL}^{cls}$, our model achieves the highest $acc_{S}$ / $acc_{U}$ scores with $77.14\%$ / $43.91\%$ compared to $56.26\%$ / $39.77\%$ achieved by AVGZSLNet / Hyper-multiple. On $\text{ActivityNet-GZSL}^{cls}$, we achieve the best $acc_{S}$ score with $45.98\%$, compared to $24.04\%$ achieved by AVCA. For the $acc_{U}$ performance, only Hyper-multiple performs better than our method with $21.30\%$ vs. $20.06\%$. 

The GZSL and ZSL results show, that when using CLIP and CLAP as feature extraction methods, a rather simple model like our method, is able to outperform methods that use more sophisticated architectural components such as cross-attention used in AVCA and Hyper-multiple, or complex concepts from hyperbolic geometry used in Hyper-multiple. 

Our method uses around $2.2$ million parameters, whereas CJME and AVGZSLNet use $2.3$ million parameters, and AVCA and Hyper-multiple have approximately $2.4$ million parameters. These numbers do not include the number of parameters of the feature extraction methods CLIP and CLAP. 

\begin{figure*}[!h]
  \centering
  \includegraphics[width=1.0\textwidth]{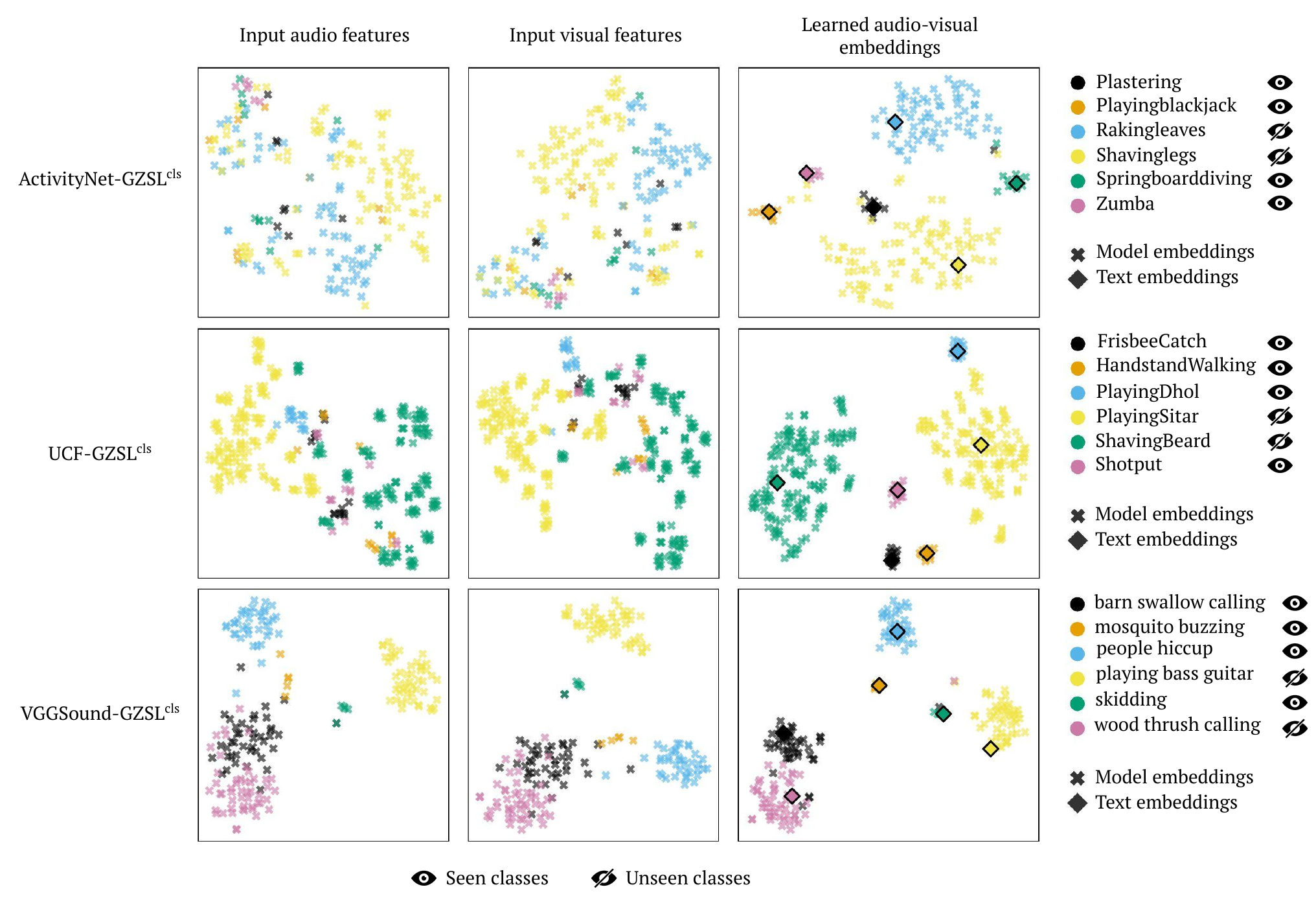}
  \caption{t-SNE visualizations for the audio input features (left), visual input features (center), and the learned output embeddings for our model (right) for the $\text{ActivityNet-GZSL}^{cls}$ (top), $\text{UCF-GZSL}^{cls}$ (center) and $\text{VGGSound-GZSL}^{cls}$ (bottom) datasets for two unseen classes and four seen classes. The learned class text embeddings are visualized as diamonds.
  }
  \label{fig:tsne_merged}
  \vspace{-0.0em}
\end{figure*}

Overall, our model achieves significant improvements over the baselines, while it requires marginally fewer parameters. Furthermore, unlike AVCA \cite{AVCA} and Hyper-multiple \cite{hyper_avca}, our method does not require positive and negative samples to calculate a triplet loss during training. This reduces memory requirements and allows for a larger batch size. \\
\mypara{Qualitative results.}
We provide t-SNE visualisations (\cref{fig:tsne_merged}) of our learned output embeddings on the $\text{VGGSound-GZSL}^{cls}$, $\text{UCF-GZSL}^{cls}$ and $\text{ActivityNet-GZSL}^{cls}$ datasets. Two unseen classes and four seen classes were randomly selected from the test set. For the unseen classes, all samples were used for the visualization, for the seen classes, all samples from the test set were used. This results in a class imbalance in the plots, since some seen classes have only a few test samples.

It can be observed that while the input features are not clustered well for all datasets, the t-SNE plots for the model outputs shows well-separated clusters for all classes.
In particular, the unseen classes are very well-separated. This shows that our method learns useful embeddings for both seen and unseen classes.
Only on $\text{VGGSound-GZSL}^{cls}$, our model does not separate well the unseen class \textit{wood thrush calling} and the seen class \textit{barn swallow calling}. This might come from the fact, that both classes can be categorized as bird sounds.
Finally, all the text embeddings are located inside the cluster of the class they belong to. This shows that our approach effectively learns to assign the audio-visual input features to the correct class.

\subsection{Ablation studies}
\label{subsec:ablations}
In this section, we conduct ablations studies on different model design choices. First, we ablate the choice of using two different textual label embeddings. Then, we study the benefit of using multi-modal inputs (audio and visual). Thirdly, we analyze the influence of the different loss components.

\setlength{\tabcolsep}{8pt} 
\renewcommand{\arraystretch}{1.1}
\begin{table*}[tp]
\centering
\begin{adjustbox}{width=\textwidth}
\begin{tabular}{l|cccc|cccc|cccc}
\hline
Label Embedding      & \multicolumn{4}{c|}{$\text{VGGSound-GZSL}^{cls}$} & \multicolumn{4}{c|}{$\text{UCF-GZSL}^{cls}$}      & \multicolumn{4}{c}{$\text{ActivityNet-GZSL}^{cls}$}                          \\
\multicolumn{1}{c|}{} & $acc_{\text{S}}$      & $acc_{\text{U}}$     & HM    & $acc_{\text{ZSL}}$  & $acc_{\text{S}}$     & $acc_{\text{U}}$     & HM    & $acc_{\text{ZSL}}$   & $acc_{\text{S}}$     & $acc_{\text{U}}$              & HM             & $acc_{\text{ZSL}}$            \\ \hline
CLIP ($w^v$)                 & 28.30  & 8.75  & 13.37 & 9.28 & 75.91 & 32.83 & 45.83 & 37.47 & 43.91 & \textbf{21.04} & \textbf{28.45} & \textbf{23.18} \\
CLAP ($w^a$)               & 18.71  & 8.94  & 12.10 & 9.09 & 53.09 & 39.62 & 45.38 & 39.78 & 35.08 & 13.03          & 19.00          & 14.20          \\
Both (Ours) &
  \textbf{29.68} &
  \textbf{11.12} &
  \textbf{16.18} &
  \textbf{11.53} &
  \textbf{77.14} &
  \textbf{43.91} &
  \textbf{55.97} &
  \textbf{46.96} &
  \textbf{45.98} &
  20.06 &
  27.93 &
  22.76 \\ \hline
\end{tabular}
\end{adjustbox}
\caption{Influence of using the two input label embeddings from CLIP and CLAP on the $\text{VGGSound-GZSL}^{cls}$, $\text{UCF-GZSL}^{cls}$ and $\text{ActivityNet-GZSL}^{cls}$ datasets.}
\label{tab:ablation_word_embeddings}
\vspace{-1.0em}
\end{table*}

\setlength{\tabcolsep}{8pt} 
\renewcommand{\arraystretch}{1.1}
\begin{table*}[tp]
\centering
\begin{adjustbox}{width=\textwidth}
\begin{tabular}{l|cccc|cccc|cccc}
\hline
Modality              & \multicolumn{4}{c|}{$\text{VGGSound-GZSL}^{cls}$} & \multicolumn{4}{c|}{$\text{UCF-GZSL}^{cls}$}      & \multicolumn{4}{c}{$\text{ActivityNet-GZSL}^{cls}$}        \\
\multicolumn{1}{c|}{} & $acc_{\text{S}}$      & $acc_{\text{U}}$     & HM    & $acc_{\text{ZSL}}$  & $acc_{\text{S}}$     & $acc_{\text{U}}$     & HM    & $acc_{\text{ZSL}}$   & $acc_{\text{S}}$     & $acc_{\text{U}}$              & HM    & $acc_{\text{ZSL}}$   \\ \hline
Audio ($a$)                 & 17.48  & 9.12  & 11.99 & 9.34 & 35.59 & 39.69 & 37.53 & 41.13 & 10.72 & 6.58           & 8.15  & 6.75  \\
Visual ($v$)               & 15.39  & 7.00  & 9.62  & 7.16 & 53.65 & 43.13 & 47.82 & 43.98 & 38.59 & \textbf{20.40} & 26.69 & 22.58 \\
Both (Ours) &
  \textbf{29.68} &
  \textbf{11.12} &
  \textbf{16.18} &
  \textbf{11.53} &
  \textbf{77.14} &
  \textbf{43.91} &
  \textbf{55.97} &
  \textbf{46.96} &
  \textbf{45.98} &
  20.06 &
  \textbf{27.93} &
  \textbf{22.76} \\ \hline
\end{tabular}
\end{adjustbox}
\caption{Influence of using only one modality or both modalities as inputs for our method on the $\text{VGGSound-GZSL}^{cls}$, $\text{UCF-GZSL}^{cls}$ and $\text{ActivityNet-GZSL}^{cls}$ datasets.}
\label{tab:ablations_modality}
\vspace{-0.0em}
\end{table*}
\setlength{\tabcolsep}{8pt}
\renewcommand{\arraystretch}{1.1}

\begin{table*}[tp]
\centering
\begin{adjustbox}{width=\textwidth}
\begin{tabular}{l|cccc|cccc|cccc}
\hline
Loss                  & \multicolumn{4}{c|}{$\text{VGGSound-GZSL}^{cls}$}                  & \multicolumn{4}{c|}{$\text{UCF-GZSL}^{cls}$}      & \multicolumn{4}{c}{$\text{ActivityNet-GZSL}^{cls}$}                        \\
\multicolumn{1}{c|}{} & $acc_{\text{S}}$     & $acc_{\text{U}}$             & HM    & $acc_{\text{ZSL}}$            & $acc_{\text{S}}$     & $acc_{\text{U}}$     & HM    & $acc_{\text{ZSL}}$   & $acc_{\text{S}}$     & $acc_{\text{U}}$              & HM            & $acc_{\text{ZSL}}$           \\ \hline
$l_{reg}$               & 5.41  & 9.44          & 6.87  & 10.03           & 22.76 & 25.49 & 24.05 & 28.22 & 5.01  & 6.69  & 5.73 & 7.11 \\
$l_{reg}+l_{ce}$       & \textbf{32.64} & \textbf{11.91} & \textbf{17.45} & \textbf{12.47} & 76.79 & 40.30 & 52.86 & 43.16 & 38.93 & \textbf{20.37} & 26.75         & 22.73         \\
$l_{reg}+l_{ce}+l_{rec}$ &
  29.68 &
  11.12 &
  16.18 &
  11.53 &
  \textbf{77.14} &
  \textbf{43.91} &
  \textbf{55.97} &
  \textbf{46.96} &
  \textbf{45.98} &
  20.06 &
  \textbf{27.93} &
  \textbf{22.76} \\ \hline
\end{tabular}
\end{adjustbox}
\caption{Influence of using different components of the loss function on the (G)ZSL performance for the $\text{VGGSound-GZSL}^{cls}$, $\text{UCF-GZSL}^{cls}$ and $\text{ActivityNet-GZSL}^{cls}$ datasets. }
\label{tab:ablation_loss}
\vspace{-1.0em}
\end{table*}

\mypara{Class label embeddings.}
Our proposed framework uses two class label embeddings as inputs. 
Here, we compare this strategy to the usage of only a single class label embedding. For this, we train our model only using CLIP text embeddings $w^v$, or only using CLAP text embeddings $w^a$. However, we use both audio and visual input features for this experiment.
The results are presented in \cref{tab:ablation_word_embeddings}. 

Using either only $w^v$ or only $w^a$ as input class label embedding leads to very similar results on $\text{VGGSound-GZSL}^{cls}$ and $\text{UCF-GZSL}^{cls}$. However, our proposed method that uses both $w^v$ and $w^a$, outperforms them significantly. 
On UCF-GZSL$^{cls}$, using both text embeddings obtains a HM 55.97\% vs.\ $45.83\%$ for $w^v$, while on VGGSound$^{cls}$ our method obtains a HM of $16.18\%$ vs.\ $13.37\%$ for $w^v$. The same trends can be observed for the ZSL performance. On $\text{ActivityNet-GZSL}^{cls}$, using $w^v$ leads to HM / $acc_{\text{ZSL}}$ scores of $28.45\%$ / $23.18\%$, while using both $w^v$ and $w^a$ performs slightly worse, achieving HM / $acc_{\text{ZSL}}$ scores of $27.93\%$ / $22.76\%$.

Finally, using both label emebeddings help significantly in terms of the $acc_{U}$ score. For VGGSound-GZSL$^{cls}$, we boost performance from $8.94\%$ for $w^a$ to $11.12\%$, while on UCF-GZSL$^{cls}$ we improve the performance from $39.62\%$ for $w^a$ to $43.91\%$ when using both label embeddings (Both). On ActivityNet-GZSL$^{cls}$, using both embeddings gives slightly lower numbers than using only $w^v$ in terms of the $acc_{U}$ scores. On the other hand, our method obtains the best $acc_{S}$ results on all three datasets.
Overall, jointly using both $w^a$ and $w^v$ provides a significant boost in performance across all the metrics and datasets.

\mypara{Multi-modality.}
In \cref{tab:ablations_modality}, we present the impact of using multi-modal input data. To obtain results for using a single input modality, only the audio or visual input feature ($a$ or $v$) along with the corresponding text embedding ($w^a$ or $w^v$) is used. 

On VGGSound-GZSL$^{cls}$, using only the audio modality achieves higher HM and $acc_{\text{ZSL}}$ scores compared to using the visual modality with HM / $acc_{\text{ZSL}}$ scores of $11.99\%$ / $9.34\%$ vs. $9.62\%$ / $7.16\%$ for the visual modality. This is likely due to VGGSound being curated specifically to include relevant audio information. In contrast, for UCF-GZSL$^{cls}$ and ActivityNet-GZSL$^{cls}$, using only the visual modality achieves better results than the audio modality on its own.
On ActivityNet-GZSL$^{cls}$, the audio modality results in HM / $acc_{\text{ZSL}}$ scores of $8.15\%$ / $6.75\%$ while using visual inputs gives HM / $acc_{\text{ZSL}}$ scores of $26.69\%$ / $22.58\%$. 

Across all datasets, the $acc_{S}$ score is significantly improved when using both modalities compared to using only $a$ or $v$. 
On UCF-GZSL$^{cls}$, our full model (Both) yields a $acc_{S}$ performance of $77.14\%$ vs. $53.65\%$ for $v$ and $35.59\%$ for $a$. For the $acc_{U}$ score, we slightly improve upon the $v$, and significantly improve over $a$. The same trend can be observed on VGGSound$^{cls}$ where our model (Both) significantly outperforms both $a$ and $v$ in $acc_{U}$ and $acc_{S}$. On ActivityNet$^{cls}$, our full model is significantly stronger in terms of the $acc_{S}$ score, while it is slightly outperformed in terms of $acc_{U}$ when using only the visual modality $v$. 

Overall, using both modalities as inputs is sound and leads to the best performance. These results highlight the fact that our model effectively exploits cross-modal relationships through the fusion of audio and visual modalities by using linear layers.

\mypara{Training objective.}
We present results for using different loss functions in \cref{tab:ablation_loss}.
Only using the regression loss $l_{reg}$ yields the poorest performance on all three datasets, with HM scores of $6.87\%$ / $24.05\%$ / $5.73\%$ for $\text{VGGSound-GZSL}^{cls}$ / $\text{UCF-GZSL}^{cls}$ / $\text{ActivityNet-GZSL}^{cls}$.
Using the cross-entropy loss $l_{ce}$ in addition to the regression loss drastically improves the performance with HM scores of $17.45 \%$ / $52.86\%$ / $26.75\%$ on $\text{VGGSound-GZSL}^{cls}$ / $\text{UCF-GZSL}^{cls}$/ $\text{ActivityNet-GZSL}^{cls}$.
Finally, adding the reconstruction loss, i.e.\ when using the full loss function $l_{reg}+l_{ce}+l_{rec}$, we achieve the best overall GZSL results.
While the impact of the reconstruction loss is smaller compared to the other two components, it still bring gains in performance. 
We observe a similar pattern for $acc_{\text{ZSL}}$.

Furthermore, on $\text{VGGSound-GZSL}^{cls}$, $acc_S$ heavily benefits from adding the cross-entropy loss function. For all three datasets, one can observe at least a three-fold improvement. Moreover, we see improvements in the $acc_U$, where $l_{ce}$ brings significant improvements. On $\text{UCF-GZSL}^{cls}$, our proposed loss function performs best in terms of both $acc_S$ and $acc_U$. On $\text{ActivityNet-GZSL}^{cls}$, the complete loss function achieves the best $acc_S$, while the $l_{reg}+l_{ce}$ loss function gives the best $acc_{U}$ scores. Finally, on $\text{VGGSound-GZSL}^{cls}$, $l_{reg}+l_{ce}$ obtains a slightly better $acc_{S}$ and $acc_{U}$ score than our full loss.
Overall, this shows that the full training objective provides the best  results across all evaluation metrics.

\section{Limitations}
\label{sec:limitations_discussion}
Our proposed method sets the new state of the art for audio-visual ZSL on three benchmark datasets when using CLIP and CLAP features. However, since the dataset used to train CLIP is not publicly available, we cannot guarantee that no unssen classes were used. Similarly, we did not attempt to remove unseen classes from the \textit{WavCaps} dataset used to train CLAP. However, \cite{mayilvahanan2023does} shows that information leakage from CLIP pre-training to image ZSL is not very significant. Incorporating CLIP encoders into the model architecture is already an established practice in current research in zero-shot / open-vocabulary semantic segmentation \cite{zhou2023zegclip, ding2022decoupling, xu2022simple, luo2023segclip, liang2023open}. 
As CLIP and CLAP were not specifically trained for the task of audio-visual GZSL, our problem setting requires significant transfer of knowledge to the new task. 

\section{Conclusion}
\label{sec:conclusion}

In this paper, we explored the usage of pre-trained large multi-modal models for audio-visual generalized zero-shot learning.
Our proposed framework ingests features extracted from the CLIP \cite{CLIP} and CLAP \cite{wavcaps} models. One of the advantages of both of the feature extraction methods is that they are also able to produce textual input embeddings for the class labels.
We proposed a simple model that consists of feed-forward neural networks and is trained with a composite loss function. When utilizing input features and both label embeddings obtained from CLIP and CLAP, our method achieves state-of-the-art results on the $\text{VGGSound-GZSL}^{cls}$, $\text{UCF-GZSL}^{cls}$, and $\text{ActivityNet-GZSL}^{cls}$ datasets.

\mypara{Acknowledgements:} This work was in part supported by BMBF FKZ: 01IS18039A, DFG: SFB 1233 TP 17 - project number 276693517, by the ERC (853489 - DEXIM), and by EXC number 2064/1 – project number 390727645. The authors thank the International Max Planck Research School for Intelligent Systems (IMPRS-IS) for supporting O.-B. Mercea.

{
    \small
    \bibliographystyle{ieeenat_fullname}
    \bibliography{main}
}

\clearpage

\setcounter{page}{1}
\maketitlesupplementary

\maketitle

\appendix

\section{Additional Details about Textual Feature Extraction}
\label{suppl:feature_extraction}

\subsection{CLIP Feature Extraction}
\label{supplement:clip_feature_extraction}
To boost zero-shot classification performance, \cite{CLIP} calculate normalized CLIP text embeddings for an ensemble of text prompts to retrieve final textual embeddings. Then the mean is taken and the result is normalized again. Normalizing the individual CLIP text representations is necessary in order to obtain a meaningful averaged vector. The second normalization facilitates the calculation of cosine similarity scores. Note that image embeddings are normalized as well. 

For $\text{UCF-GZSL}^{cls}$ and $\text{ActivityNet-GZSL}^{cls}$, we use an ensemble of $48$ different prompt templates for each class. $\text{UCF-GZSL}^{cls}$ and $\text{ActivityNet-GZSL}^{cls}$ have a similar context since both are action recognition datasets. Hence, we use the same text prompts for these two datasets (see listing \ref{app:clip_text_prompts}). These templates are taken from the CLIP repository\footnote{\href{https://github.com/openai/CLIP/blob/main/data/prompts.md\#ucf101}{https://github.com/openai/CLIP/blob/main/data/prompts.md\#ucf101}}. 

\begin{lstlisting}[caption=Text prompt templates that were used to create CLIP label embeddings for $\text{UCF-GZSL}^{cls}$ and $\text{ActivityNet-GZSL}^{cls}$., label={app:clip_text_prompts}, language=Python]
CLIP_prompt_templates = [
    'a photo of a person {}.',
    'a video of a person {}.',
    'a example of a person {}.',
    'a demonstration of a person {}.',
    'a photo of the person {}.',
    'a video of the person {}.',
    'a example of the person {}.',
    'a demonstration of the person {}.',
    'a photo of a person using {}.',
    'a video of a person using {}.',
    'a example of a person using {}.',
    'a demonstration of a person using {}.',
    'a photo of the person using {}.',
    'a video of the person using {}.',
    'a example of the person using {}.',
    'a demonstration of the person using {}.',
    'a photo of a person doing {}.',
    'a video of a person doing {}.',
    'a example of a person doing {}.',
    'a demonstration of a person doing {}.',
    'a photo of the person doing {}.',
    'a video of the person doing {}.',
    'a example of the person doing {}.',
    'a demonstration of the person doing {}.',
    'a photo of a person during {}.',
    'a video of a person during {}.',
    'a example of a person during {}.',
    'a demonstration of a person during {}.',
    'a photo of the person during {}.',
    'a video of the person during {}.',
    'a example of the person during {}.',
    'a demonstration of the person during {}.',
    'a photo of a person performing {}.',
    'a video of a person performing {}.',
    'a example of a person performing {}.',
    'a demonstration of a person performing {}.',
    'a photo of the person performing {}.',
    'a video of the person performing {}.',
    'a example of the person performing {}.',
    'a demonstration of the person performing {}.',
    'a photo of a person practicing {}.',
    'a video of a person practicing {}.',
    'a example of a person practicing {}.',
    'a demonstration of a person practicing {}.',
    'a photo of the person practicing {}.',
    'a video of the person practicing {}.',
    'a example of the person practicing {}.',
    'a demonstration of the person practicing {}.',
]
\end{lstlisting}

$\text{VGGSound-GZSL}^{cls}$ contains videos of a variety of categories and hence more general prompts are required. The prompts that we used to create CLIP text embeddings for $\text{VGGSound-GZSL}^{cls}$ can be seen in listing \ref{listing:vggsound_clip_text_prompts}.

\begin{lstlisting}[caption=Text prompt templates that were used to create CLIP text embeddings for $\text{VGGSound-GZSL}^{cls}$., label={listing:vggsound_clip_text_prompts}, language=Python]
VGGSound_CLIP_prompt_templates = [
    'a photo of {}.',
    'a video of {}.',
    'a example of {}.',
    'a demonstration of {}.',
    'a photo of the person {}.',
    'a video of the {}.',
    'a example of the {}.',
    'a demonstration of the {}.'
]
\end{lstlisting}

\subsection{CLAP Feature Extraction}
\label{supplement:clap_feature_extraction}

We use the same procedure as in \ref{supplement:clip_feature_extraction} to extract textual CLAP embeddings. For $\text{UCF-GZSL}^{cls}$ and $\text{ActivityNet-GZSL}^{cls}$ we use the prompts as in listing \ref{app:wavcaps_text_prompts}. For $\text{VGGSound-GZSL}^{cls}$, we use the prompts given in listing \ref{listing:vggsound_clap_text_prompts}.

\begin{lstlisting}[caption=Text prompt templates that were used to create CLAP label embeddings for $\text{UCF-GZSL}^{cls}$ and $\text{ActivityNet-GZSL}^{cls}$., label={app:wavcaps_text_prompts}, language=Python]
CLAP_prompt_templates = [
    'a person {} can be heard.',
    'a example of a person {} can be heard.',
    'a demonstration of a person {} can be heard.',
    'the person {} can be heard.',
    'a example of the person {} can be heard.',
    'a demonstration of the person {} can be heard.',
    'a person using {} can be heard.',
    'a example of a person using {} can be heard.',
    'a demonstration of a person using {} can be heard.',
    'a example of the person using {} can be heard.',
    'a demonstration of the person using {} can be heard.',
    'a person doing {} can be heard.',
    'a example of a person doing {} can be heard.',
    'a demonstration of a person doing {} can be heard.',
    'a example of the person doing {} can be heard.',
    'a demonstration of the person doing {} can be heard.',
    'a example of a person during {} can be heard.',
    'a demonstration of a person during {} can be heard.',
    'a example of the person during {} can be heard.',
    'a demonstration of the person during {} can be heard.',
    'a person performing {} can be heard.',
    'a example of a person performing {} can be heard.',
    'a demonstration of a person performing {} can be heard.',
    'a example of the person performing {} can be heard.',
    'a demonstration of the person performing {} can be heard.',
    'a person practicing {} can be heard.',
    'a example of a person practicing {} can be heard.',
    'a demonstration of a person practicing {} can be heard.',
    'a example of the person practicing {} can be heard.',
    'a demonstration of the person practicing {} can be heard.'
]
\end{lstlisting}

\begin{lstlisting}[caption=Text prompt templates that were used to create CLAP text embeddings for $\text{VGGSound-GZSL}^{cls}$., label={listing:vggsound_clap_text_prompts}, language=Python]
VGGSound_CLAP_prompt_templates = [
    'a {} can be heard.',
    'a example of a {} can be heard.',
    'a demonstration of a {} can be heard.',
    'the {} can be heard.',
    'a example of the {} can be heard.',
    'a demonstration of the {} can be heard.',
    '{} can be heard.',
    'a example of {} can be heard.',
    'a demonstration of {} can be heard.'
]
\end{lstlisting}

\end{document}